\documentclass[10pt,twocolumn,letterpaper]{article}

\usepackage{iccv}
\usepackage{times}
\usepackage{epsfig}
\usepackage{graphicx}
\usepackage{amsmath}
\usepackage{amssymb}
\usepackage{booktabs}
\usepackage{appendix}

\DeclareMathOperator{\EX}{\mathbb{E}} 

\usepackage[pagebackref=true,breaklinks=true,letterpaper=true,colorlinks,bookmarks=false]{hyperref}

\iccvfinalcopy 

\ificcvfinal\fi

\title{Compositional GAN: Learning Image-Conditional Binary Composition}

\author{
Samaneh Azadi, Deepak Pathak, Sayna Ebrahimi, Trevor Darrell\\
  University of California, Berkeley\\
  \texttt{\{sazadi, pathak, sayna, trevor\}@eecs.berkeley.edu} \\
}

\begin{document}
\setlength{\abovedisplayskip}{4pt}
\setlength{\belowdisplayskip}{3pt}
\setlength{\belowcaptionskip}{-10pt}

\maketitle

\begin{abstract}
Generative Adversarial Networks (GANs) can produce images of remarkable complexity and realism but are generally structured to sample from a single latent source ignoring the explicit spatial interaction between multiple entities that could be present in a scene. Capturing such complex interactions between different objects in the world, including their relative scaling, spatial layout, occlusion, or viewpoint transformation is a challenging problem. In this work, we propose a novel self-consistent Composition-by-Decomposition (CoDe) network to compose a pair of objects. Given object images from two distinct distributions, our model can generate a realistic composite image from their joint distribution following the texture and shape of the input objects. We evaluate our approach through qualitative experiments and user evaluations. Our results indicate that the learned model captures potential interactions between the two object domains, and generates realistic composed scenes at test time.
\end{abstract}

\section{Introduction}
Conditional Generative Adversarial Networks (cGANs) have emerged as a powerful method for generating images conditioned on a given input. The input cue could be in the form of an image~\cite{isola2017image,zhu2017unpaired, liu2017unsupervised, azadi2017multi,wang2017high,pathakCVPR16context}, a text phrase~\cite{zhang2016stackgan,reed2016learning, reed2016generative,johnson2018image} or a class label layout~\cite{mirza2014conditional,odena2016conditional,antoniou2017data}. The goal in most of these GAN instantiations is to learn a mapping that \textit{translates} a given sample from the source distribution to generate a sample from the output distribution. This primarily involves transforming either a single object of interest (apples to oranges, horses to zebras, label to image, etc.) or changing the style and texture of the input image (day to night, etc.). However, these direct transformations do not capture the fact that a natural image is a 2D projection of a \textit{composition} of multiple objects interacting in a 3D visual world. Here, we explore the role of compositionality in GAN frameworks and propose a new method which learns a function that maps images of different objects sampled from their marginal distributions (e.g., chair and table) into a combined sample (table-chair) that captures the joint distribution of object pairs. In this paper, we specifically focus on the composition of a pair of objects.

\begin{figure}[t]
\centering
\includegraphics[width=0.5\textwidth]{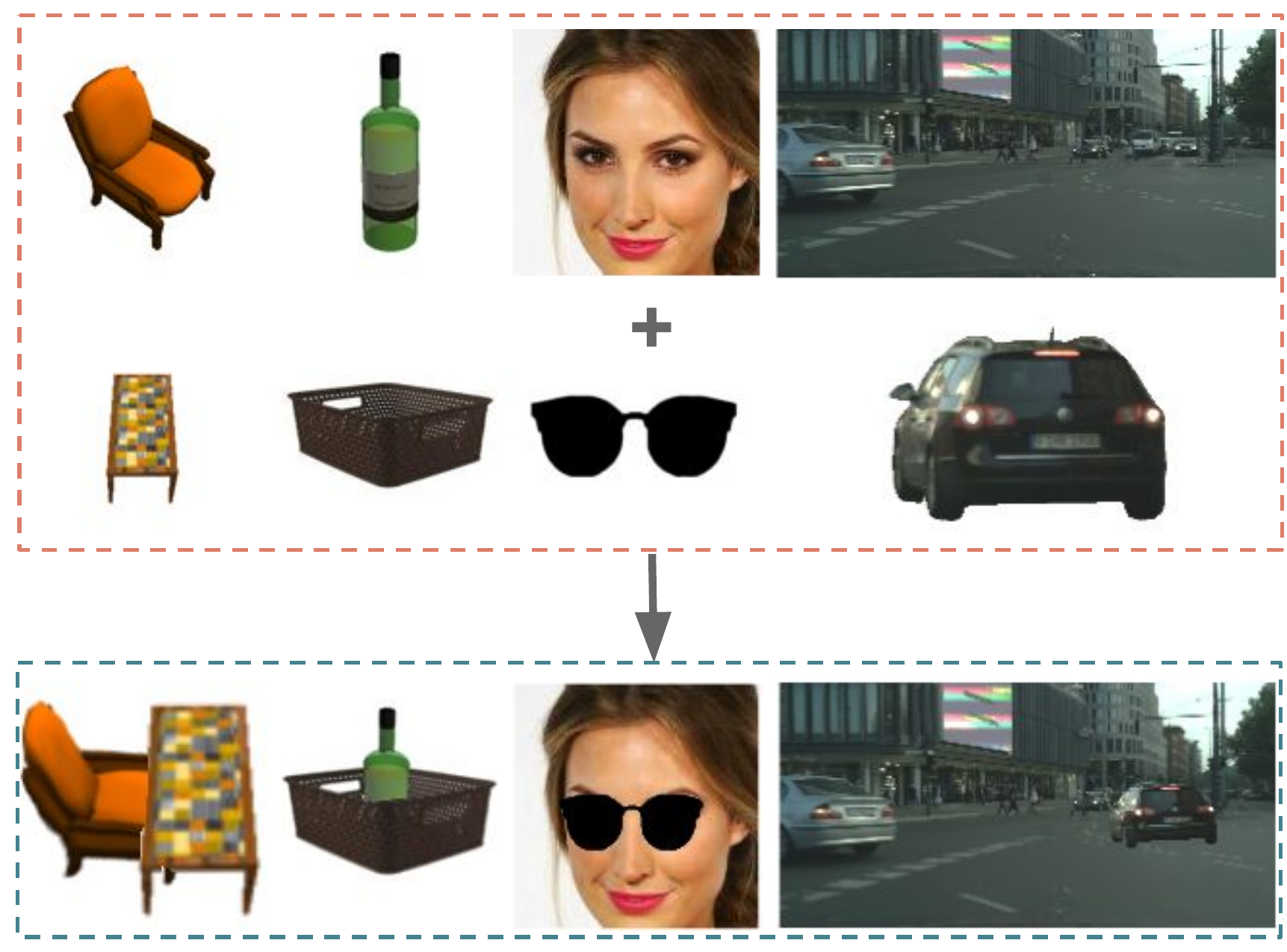}
\caption{Binary Composition examples. \textit{Top Row}: The first object or the background image, \textit{Middle Row}: The second object or the foreground image, \textit{Bottom Row}: The generated composite image.} 
\label{fig:teaser}
\end{figure}

Modeling compositionality in natural images is a challenging problem due to the complex interactions among different objects with respect to relative scaling, spatial layout, occlusion or viewpoint transformation. Recent work using spatial transformer networks~\cite{spatial2015} within a GAN framework~\cite{lin2018st} decomposes this problem by operating in a geometric warp parameter space to find a geometric modification for a foreground object. However, this approach is only limited to a fixed background and does not consider more complex interactions in the real world. 

We consider the task of composing two input object images into  a joint image that captures their realistic interactions. For instance, given an image of a chair and an image of a table, our formulation is able to generate an image containing the same chair-table pair arranged in a realistic manner. To the best of our knowledge, this is the first work that addresses the problem of generating a composed image from two given inputs using a GAN, trainable under paired and unpaired scenarios. In an unpaired training setup, one does not have access to the paired examples of the same object instances with their combined compositional image. For instance, to generate the joint image from the image of a given table and a chair, we might not have any example of that particular chair beside that particular table while we might have images of other chairs and other tables together. 

Our key insight is to leverage the idea that a successful composite image of two objects should not only be realistic in appearance but can also be decomposed back into individual objects. Hence, we use decomposition as a supervisory signal to train composition, thereby enforcing a self-consistency constraint~\cite{zhu2017unpaired} through a composition-by-decomposition (CoDe) network.
Moreover, we use this self-consistent CoDe network for an example-specific meta-refinement (ESMR) approach at test time to generate sharper and more accurate composite images: We fine-tune the weights of the composition network on each given test example by the self-supervision provided from the decomposition network.

Through qualitative and quantitative experiments, we evaluate our proposed Compositional-GAN approach in two training scenarios: (a) paired: when we have access to paired examples of individual object images with their corresponding composed image, (b) unpaired: when we have a dataset from the joint distribution without being paired with any of the images from the marginal distributions.
Our compositional GAN code and dataset will be available at \url{https://github.com/azadis/CompositionalGAN}.

\section{Related Work}
\textbf{Generative adversarial networks (GANs)} have been used in a wide variety of settings including image generation~\cite{denton2015deep,yang2017lr, karras2017progressive} and representation learning~\cite{radford2015unsupervised,salimans2016improved,liu2016coupled,xi2016infogan}. The loss function in GANs has been shown to be effective in optimizing high quality images conditioned on an available information. Conditional GANs~\cite{mirza2014conditional} generate appealing images in a variety of applications including image to image translation both in the case of paired~\cite{isola2017image} and unpaired data~\cite{zhu2017unpaired, zhu2017toward}, inpainting missing image regions~\cite{pathakCVPR16context,yang2016high}, generating photorealistic images from labels~\cite{mirza2014conditional,odena2016conditional}, and solving for photo super-resolution~\cite{ledig2016photo,lai2017deep}.

\textbf{Image composition} is a challenging problem in computer graphics where objects from different images are to be overlaid in one single image. The appearance and geometric differences between these objects are the obstacles that can result in non-realistic composed images. \cite{zhu2015learning} addressed the composition problem by training a discriminator that could distinguish realistic composite images from synthetic ones. \cite{tsai2017deep} developed an end-to-end deep CNN for image harmonization to automatically capture the context and semantic information of the composite image. This model outperformed its precedents \cite{sunkavalli2010multi,xue2012understanding} which transferred statistics of hand-crafted features to harmonize the foreground and the background in the composite image. Recently, \cite{lin2018st} used spatial transformer networks as a generator by performing geometric corrections to warp a masked object to adapt to a fixed background image. Moreover, ~\cite{johnson2018image} computed a scene layout from given scene graphs which revealed explicit reasoning about relationships between objects and converted the layout to an output image. In the image-conditional composition problem which we address, each object should be rotated, scaled, and translated while partially occluding the other object to generate a realistic composite image.   

\section{Background: Conditional GAN}
\label{sec:background}

We briefly review the conditional Generative Adversarial Networks before discussing our compositional setup. Given a random noise vector $z$, GANs generate images $c$ of a specific distribution using a generator $G$ which is trained adversarially with respect to a discriminator $D$.
While the generator tries to produce realistic images, the discriminator opposes the generator by learning to distinguish between real and fake images. In conditional GAN models (cGANs), an auxiliary information $x$ is fed into the model, in the form of an image or a label, alongside the noise vector, i.e., $\{x,z\} \rightarrow c$~\cite{goodfellow2016nips,mirza2014conditional}. The objective of cGANs would be therefore an adversarial loss function formulated as
\begin{eqnarray*}
\lefteqn{\mathcal{L}_{cGAN}(G,D) =  \EX_{{x,c}\sim p_{\text{data}}(x,c)} [\log D(x,c)]} \\
&+&\EX_{x \sim p_{\text{data}}(x), z\sim p_z(z)} [1 - \log D(x,G(x,z))] \text{, where}
\end{eqnarray*}  $G$, $D$ minimize and maximize this objective, respectively. 

The convergence of the above GAN objective and consequently the quality of the generated images would be improved if an $L_1$ loss penalizing deviation of the generated images from their ground-truth is added. Thus, the generator's objective function would be summarized as
\begin{eqnarray*}
G^* &=& \arg \min_{G} \max_{D} \mathcal{L}_{cGAN}(G,D) \\
&+& \lambda \EX_{x,c\sim p_{\text{data}}(x,c), z\sim p_z(z)} [\| c - G(x,z) \|_{1}] 
\end{eqnarray*}

\begin{figure*}[t]
\centering
\includegraphics[width=0.95\textwidth]{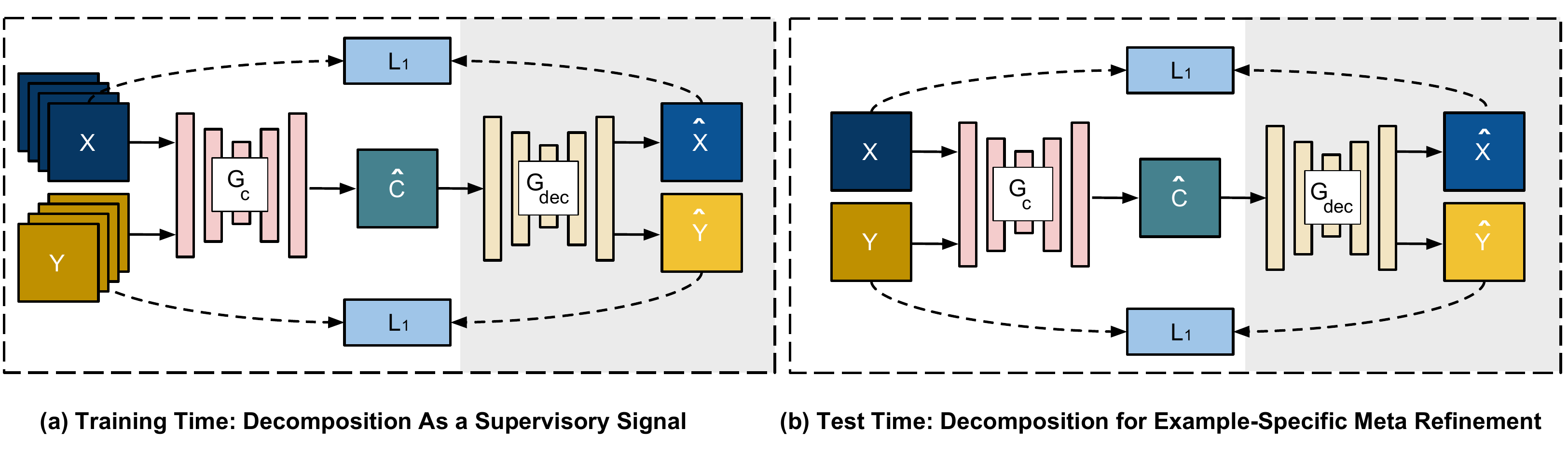}
\caption{(a) The CoDe training network includes the composition network getting a self-consistent supervisory signal from the decomposition network. This network is trained on all training images, (b) ESMR: At test time, the weights of the trained composition and decomposition networks are fine-tuned given only one test example of $X$ and one test example of $Y$. The decomposition network provides the self-supervision required for updating the weights of the composition network at test time. The layers of the composition generator are presented in pink and the decomposition generator in yellow.} 
\label{fig:comp-decomp}
\end{figure*}

\section{Compositional GAN}
\label{sec:compGAN}
Conditional GANs, discussed in Section~\ref{sec:background}, have been applied to several image translation problems such as day to night, horse to zebra, and sketch to portraint ~\cite{isola2017image, zhu2017unpaired}. However, the composition problem is more challenging than just translating images from one domain to another because the model additionally needs to handle the relative scaling, spatial layout, occlusion, and viewpoint transformation of the individual objects to generate a realistic composite image.
Here, we propose Compositional GAN for generating a composite image given two individual object images.

Let $x$ be an image containing the first object, $y$ be an image of the second object and $c$ be the image from their joint distribution. During training, we are given datasets $X = \{x_1, \cdots, x_n \}$ and $Y = \{y_1, \cdots, y_n \}$ from the marginal distribution of the two objects, and $C = \{c_1, \cdots, c_n\}$ from their joint distribution containing both objects.
We further assume that the segmentation masks of objects are available for both individual images in $X,Y$ as well as the composite images in $C$.
Our proposed binary compositional GAN model is conditioned on \textit{two} input images ($x, y$) to generate an image from the target distribution $p_{\text{data}}(c)$.
The goal is to ensure that the generated composite image $\hat{c}$ contains the objects in $x,y$ with the same color, texture, and structure while also looking realistic with respect to set $C$.
Note that instead of learning a generative model of all possible compositions, our aim is to learn a mode of the distribution.

Since the conditional GANs are not adequate for transforming objects spatially, 
we explicitly model the scale and shift transformations by translating the object images $(x, y)$ to $(x^T, y^T)$ based on a relative Spatial Transformer Network (STN) ~\cite{spatial2015}. Moreover, in specific domains where a viewpoint transformation of the first object relative to the second one is required, we propose a Relative Appearance Flow Network (RAFN). Details of our relative STN and RAFN are provided in Sections~\ref{sec:STN}, \ref{sec:AFN}. 

\subsection{Supervising composition by decomposition}
\label{comp-decomp}

The central idea of our approach is to supervise the composition of two images $(x^T$, $y^T)$ via a self-consistency loss function ensuring that the generated composite image, $\hat{c}$, can further be decomposed back into the respective individual object images.  
The composition is performed using a conditional GAN, $(G_\text{c}, D_\text{c})$, that takes the two RGB images $(x^T, y^T)$ concatenated channel-wise as the input to generate the corresponding composite output, $\hat{c}$, with the two input images appropriately composed. This generated image will then be fed into a another conditional GAN, $(G_{\text{dec}},D_{\text{dec}})$, to be decomposed back into its constituent objects, $(\hat{x}^T$, ${\hat{y}^T})$ using a self-consistency $L_1$ loss function. The schematic of our self-consistent Composition-by-Decomposition (CoDe) network is illustrated in Figure~\ref{fig:comp-decomp}-(a).

In addition to the decomposition network, the generated composite image will be given to a mask prediction network, $G^M_\text{dec}$, that predicts the probability of each pixel in the composite image to belong to each of the input objects or background. The argmax of these probabilities over object ids results in the estimated masks of the two objects as $\hat{M}_{x}$ and $\hat{M}_{y}$. The two decomposition generators $G_\text{dec}$ and $G^M_\text{dec}$ share their weights in their encoder network but are different in the decoder. A GAN loss with a gradient penalty~\cite{gulrajani2017improved} is applied on top of the generated images $\hat{c}, \hat{x}^T, \hat{y}^T $ to make them look realistic in addition to multiple $L_1$ loss functions penalizing deviation of the generated images from their ground-truth. Further details of our training model are provided in Section~\ref{sec:full}.

\noindent\textbf{Extension to Unpaired Data:}
We train our Compositional GAN framework in two scenarios: (1) when inputs-output are \textit{paired} in the training set, i.e., each composite image in $C$ has corresponding individual object images in $X, Y$, and (2) when training data is \textit{unpaired}, i.e., images in $C$ do not correspond to images in $X$ and $Y$.
We convert the unpaired data to a paired one by cutting out the respective object segments from each composite image in $C$ to get the corresponding paired individual object images.
Although these new object cutouts would be paired with the composite image, they are incomplete and not amodal because of occlusion in the composite image. Hence, we synthesize the missing part of these individual object cutouts 
using self-supervised inpainting networks~\cite{pathakCVPR16context} which are trained on object images from $X$ and $Y$, described in Section~\ref{inpainting}.

\subsection{Example-Specific Meta-Refinement (ESMR)}
\label{sec:finetune}
The compositional GAN model not only should learn to compose two object with each other, but it also needs to preserve the color, texture and other properties of the individual objects in the composite image. While our framework is able to handle the former, it suffers at times to preserve color and texture of held-out objects at test time. We propose to handle this issue by performing per-example refinement at test time. Since our training algorithm gets supervision by decomposing the composite image back into individual objects, we can use the same supervisory signal to refine the generated composite image $\hat{c}$ for unseen test examples as well. Hence, we continue to optimize the network parameters using the decomposition of the generated image back into the two test objects to remove artifacts and generate sharper results. This example-specific meta-refinement (ESMR), depicted in Figure~\ref{fig:comp-decomp}-(b), improves the quality of the composite image at inference.

Given the segmentation masks of the input object images, we again ignore background for simplicity. We freeze the weights of the relative STN, RAFN, and $G^M_\text{dec}$, while only refining the weights of the CoDe layers. A GAN loss is applied on the outputs of the generators given the real samples from our training set.  
The objective function for our ESMR approach would be thus summarized as
\begin{eqnarray}
\mathcal{L}(G) &=& \lambda (\| \hat{x}^T- x^T\|_1 + \|\hat{M}_x \odot \hat{c} - \hat{M}_x \odot x^T\|_1 \nonumber\\
&+& \| \hat{y}^T- y^T\|_1 + \|\hat{M}_y \odot \hat{c} - \hat{M}_y \odot y^T\|_1) \nonumber\\
&+& [\mathcal{L}_{\text{cGAN}}(G_c,D_c) +\mathcal{L}_{\text{cGAN}}(G_\text{dec},D_\text{dec}) ],\nonumber
\end{eqnarray} where $\hat{x}^T, \hat{y}^T$ are the generated decomposed images, and $x^T$, $y^T$ are the transposed inputs. Here, the decomposition and mask prediction networks reinforce each other in generating sharper outputs and predicting more accurate segmentation masks. 
The mask prediction loss (i.e., second and fourth $L_1$ loss terms) provides an extra supervision for occlusion ordering of the two objects in the composite image during meta-refinement optimization at inference. We quantify it through an ablation study in Section 2.3 of the Supplemental where eliminating the mask prediction network results in an incorrect occlusion ordering.
On the other hand, ignoring the self-consistency $L_1$ loss from the decomposition network (i.e., first and third loss terms) results in a composite image with the object shapes deviated from their corresponding inputs, as shown in the ablation study.

\section{Implementation Details}
In this section, we provide more details on the components of our training network including the relative STN, RAFN, inpainting, and our full end-to-end model.

\subsection{Relative spatial transformer network}
\label{sec:STN}
Given the segmentation masks of the objects for images in sets $X, Y$, and $C$, we crop and scale all input objects to be at the center of the image in all training images. To \textit{relatively} translate the center-oriented input objects, ($x$, $y$), to an appropriate spatial layout, we train our variant of the spatial transformer network (STN)~\cite{spatial2015}.
This Relative STN simultaneously takes the \textit{two} RGB images concatenated channel-wise and translates them relatively to each other to $(x^T, y^T)$ based on their spatial relation encoded in the training composite images.\footnote{\label{note1}The architecture of this network is illustrated in the Appendix.}

\subsection{Relative Appearance Flow Network (RAFN)}
\label{sec:AFN}
In specific domains where the relative viewpoint of the objects should be changed accordingly to generate a natural composite image, we introduce our relative appearance flow network orthogonal to our main CoDe pipeline. Irrespective of the paired or unpaired training data, we propose a relative encoder-decoder appearance flow network, $G_\text{RAFN}$, based on the AFN model introduced in~\cite{zhou2016view}. The AFN model of~\cite{zhou2016view} uses an explicit rotating angle parameter for synthesizing images in their target view. However, our RAFN model synthesizes a new viewpoint of the first object, $x$, given the viewpoint of the second one, $y$, \textit{encoded in its binary mask}. Our RAFN is trained on a set of images in $X$ with arbitrary azimuth angles $\alpha \in \{-180^{\circ}, -170^{\circ}, \cdots, 180^{\circ} \}$ along with their target images with arbitrary new azimuth angles $\theta \in \{-180^{\circ}, -170^{\circ}, \cdots, 180^{\circ} \}$ and a set of foreground masks of images in $Y$ in the same target viewpoints. The architecture of our RAFN is illustrated in the Appendix, and its loss function is formulated as
\begin{eqnarray}
\lefteqn{\mathcal{L}(G_{\text{RAFN}}) = \mathcal{L}_{L_1}(G_\text{RAFN}) + \lambda \mathcal{L}_\text{BCE}(G^M_\text{RAFN})}\nonumber\\
&=& \mathbb{E}_{(x,y)}[\|x -G_{\text{RAFN}}(M_{y}^{\text{fg}},x^r)\|_1] \\
&+&  \lambda \mathbb{E}_{x} [\hat{M}_{x}^{\text{fg}}\log M_{x}^{\text{fg}} + (1-\hat{M}_{x}^{\text{fg}})\log (1-M_{x}^{\text{fg}})]\nonumber
\end{eqnarray}

\begin{figure*}[t]
\centering
\includegraphics[width=\textwidth]{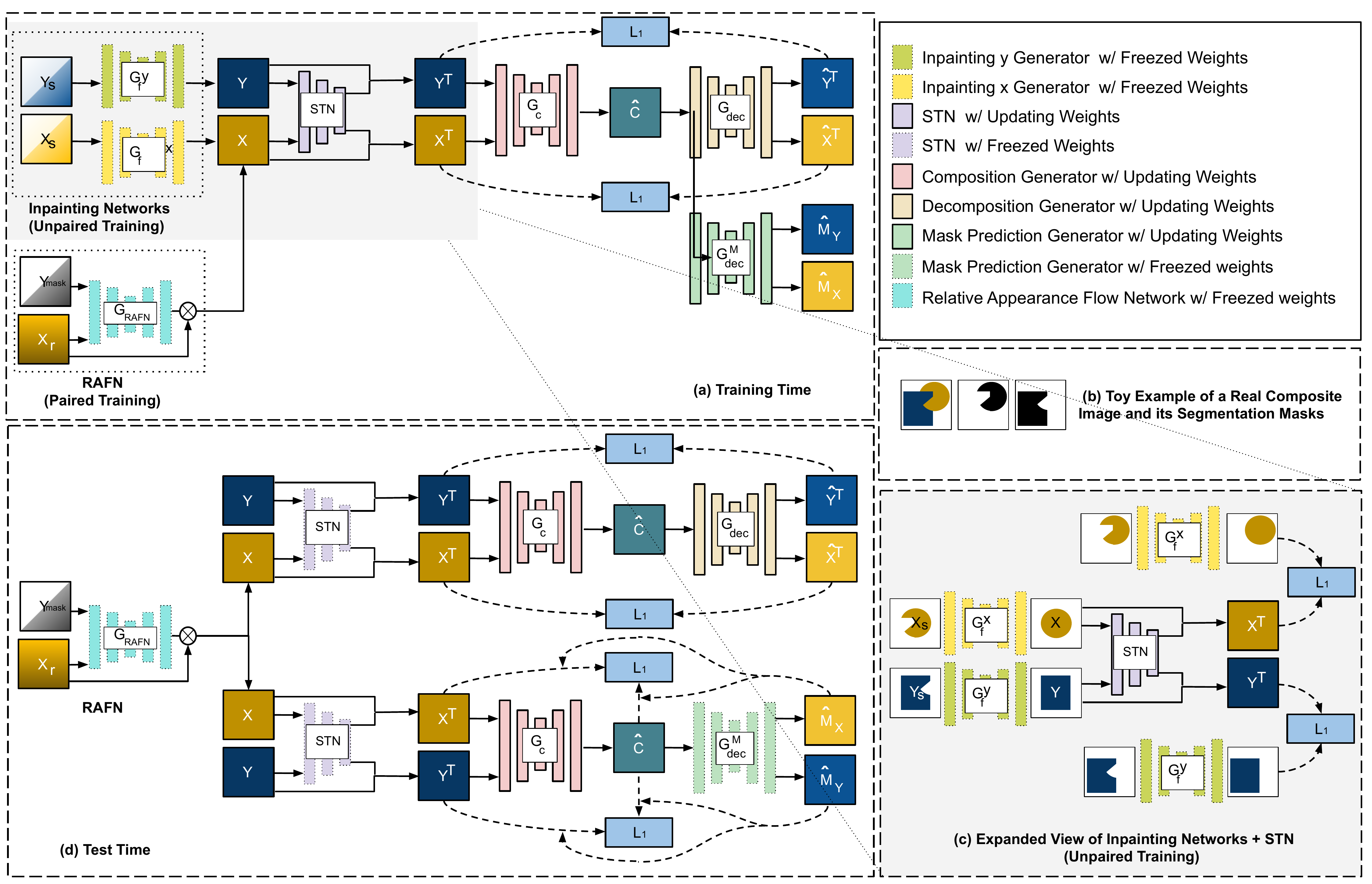}
\caption{Schematic of our binary compositional GAN model at training and test times: (a) Our training model includes the inpainting networks (for unpaired data), RAFN (for paired data), the relative STN model, and the CoDe and mask prediction networks. $X_s$ and $Y_s$ stand for the respective object segments of real composite images in an unpaired training setup. $X_r$ indicates the input image in an arbitrary viewpoint different from its corresponding composite image, and $Y_\text{mask}$ is the binary segmentation mask of the object images in $Y$ encoding the target viewpoint, (b) A toy example of a real composite image and its segmentation masks, (c) We convert an unpaired training data to a paired setup by inpainting the object segment cutouts of the real composite image. The inpainted segments and their cropped variants at the center of the image are then used for training STN, (d) At test time, we fine-tune the weights of the CoDe network given only one test example from the $X$ domain and one test example from the $Y$ domain. The weights of the mask prediction network and STN are not updated on test examples. Each of the above modules is represented by a different color, and repeating the same module in different parts of this diagram is for the illustration purpose.} 
\label{fig:train}
\end{figure*} 
Here, $G_{\text{RAFN}}$ is the encoder-decoder network predicting the appearance flow vectors, which after a bilinear sampling step generates the synthesized view. Also, $G^M_\text{RAFN}$ is an encoder-decoder mask prediction network sharing the weights of its encoder with $G_{\text{RAFN}}$, while its decoder is designed for predicting the foreground masks of the synthesized images. Moreover, $x$ is the ground-truth image for the first object in the target viewpoint while $x^r$ indicates the input image in an arbitrary view. The predicted foreground mask of RAFN is represneted by $\hat{M}_{x}^{\text{fg}}$, while $M_{x}^{\text{fg}}$, $M_{y}^{\text{fg}}$ are the ground-truth segmentation masks for objects in $x, y$, respectively\textsuperscript{\ref{note1}}.

\subsection{Inpainting network}
\label{inpainting}

\begin{figure*}[t!]
\centering
\includegraphics[width=0.9\textwidth]{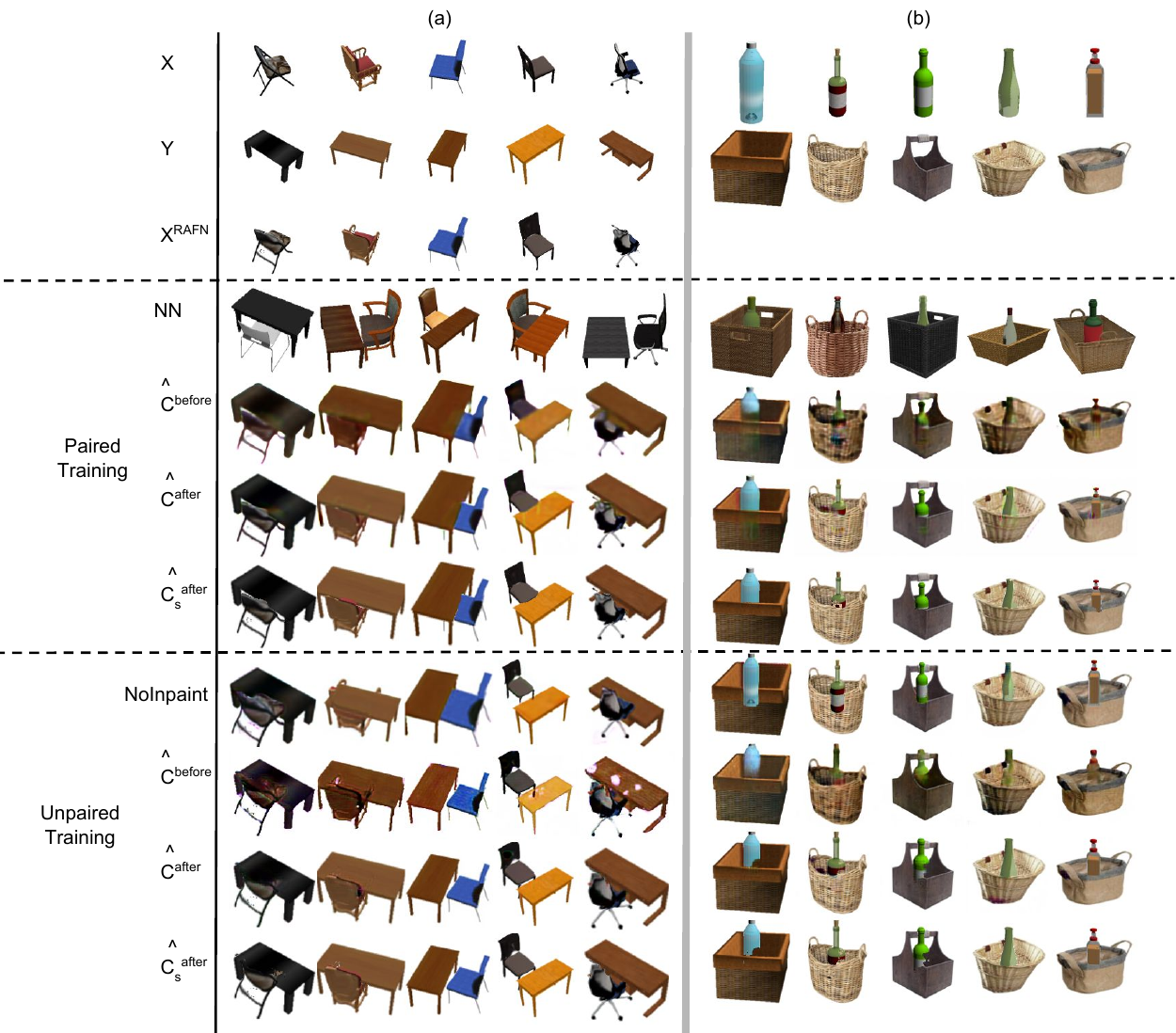}
\caption{Test results on (a) the chair-table and (b) the basket-bottle composition tasks trained with either paired  or unpaired data. ``NN'' stands for the nearest neighbor image in the paired training set, and ``NoInpaint'' shows the results of the unpaired model without the inpainting network. In both paired and unpaired cases, $\hat{c}^{\text{before}}$ and $\hat{c}^{\text{after}}$ show outputs of the generator before and after ESMR, respectively. Also, $\hat{c}^{\text{after}}_s$ represents summation of masked transposed inputs after the ESMR step.} 
\vspace{-1mm}
\label{fig:chair-table-basket-bottle}
\end{figure*}

As discussed in Section~\ref{comp-decomp}, when a paired training data is not available, we provide inputs-output pairs by inpainting the object segment cutouts of each real composite image.  We train a self-supervised inpainting network~\cite{pathakCVPR16context}, $G_\text{f}$, for each input domain to generate complete objects from the respective object segments of the composite image in $C$. A toy example is depicted in Figure~\ref{fig:train}-(c). Using inpainted segments instead of the occluded ones to be paired with the real composite image in $C$ reinforces the CoDe model in learning object occlusions and spatial layouts more accurately, as shown in Figure~\ref{fig:chair-table-basket-bottle}. 

To train the inpainting network for domain $X$ in a self-supervised manner, we apply a random binary mask on each image $x$ to zero out the pixel values inside the applied mask region. Now given the masked image and the original image $x$, we train a conditional GAN, $(G_\text{f}^x, D_\text{f}^x)$, to fill in the masked regions of the image. Here, we use the segmentation masks of object images in $Y$ as the binary masks for zeroing out the pixel values of images in $X$ partially. Another cGAN network, $(G_\text{f}^y, D_\text{f}^y)$, would be trained similarly to fill in the masked regions of images in $Y$. The loss function for each inpainting network would be:
\begin{eqnarray}
\mathcal{L}(G_\text{f}) = \mathcal{L}_{L_1}(G_\text{f}) + \lambda \mathcal{L}_{\text{cGAN}} (G_\text{f}, D_\text{f})	
\end{eqnarray}

In short, starting from the two inpainting networks trained on two sets $X$ and $Y$, we convert the unpaired data to a paired setup by generating complete objects from the object cutouts of images in $C$. This allows training our model with an unpaired training data similar to the paired case. This conversion is summarized in Figure~\ref{fig:train}-(a, c).

\subsection{Full model}
\label{sec:full}
 An schematic of our full network, $G$, is represented in Figure~\ref{fig:train} and the objective function is composed of:

\begin{itemize}
   \item  Pixel-reconstruction $L_1$-loss functions on the outputs of the composition generator, decomposition generator, and the relative STN model:
 \begin{eqnarray}
 \mathcal{L}_{L_1} (G_\text{c}) &=& \mathbb{E}_{(x, y, c)}\big[\|c - \hat{c}\|_1\big], \nonumber\\
\mathcal{L}_{L_1} (G_\text{dec}) &=&  \mathbb{E}_{(x, y)}\big[\|(x^T, y^T) - G_\text{dec}(\hat{c})\|_1\big], \nonumber\\
\mathcal{L}_{L_1} (\text{STN}) &=& \mathbb{E}_{(x, y)}\big[\|(x^c,y^c) - (x^T, y^T)\|\big]\nonumber,
 \end{eqnarray} where $(x^T, y^T) = \text{STN}(x, y)$ and $\hat{c} = G_\text{c}(x^T, y^T)$. Moreover, ($x^c$, $y^c$) are the ground-truth transposed input objects corresponding to the composite image, $c$, in the paired scenario (or equivalently, the inpainted object segments of $c$ in the unpaired case). 
 Also, $\mathcal{L}_{L_1}(G_\text{dec})$ indicates the self-consistency constraint penalizing deviation of decomposed images from their corresponding inputs,
 \item A cross-entropy mask prediction loss as  $\mathcal{L}_\text{CE}(G^M_\text{dec})$ to assign a label to each pixel of the generated composite image, $\hat{c}$, corresponding with the $\{x, y, \text{background}\}$ classes,
 \item Conditional GAN loss functions for both the composition and decomposition networks:
 \begin{eqnarray}
 \mathcal{L}_\text{cGAN} (G_{\text{c}},D_{\text{c}}) & =& \mathbb{E}_{(x, y, c)} \big[\log D_\text{c}(x^T, y^T, c)\big]\nonumber \\
&+& \mathbb{E}_{(x, y)}\big[1-\log D_\text{c}(x^T, y^T, \hat{c})\big],\nonumber
 \end{eqnarray}
 \begin{eqnarray}
 \lefteqn{\mathcal{L}_\text{cGAN} (G_{\text{dec}},D_{\text{dec}}) = \mathbb{E}_{(x, y)} \big[\log D_\text{dec}(\hat{c}, x^c)}\nonumber\\
 &+ \log D_\text{dec}(\hat{c}, y^c)\big] + \mathbb{E}_{(x, y)}\big[(1-\log D_\text{dec}(\hat{c}, \hat{x}^T)) &\nonumber\\
&+(1-\log D_\text{dec}(\hat{c}, \hat{y}^T))\big].&\nonumber
 \end{eqnarray}
 \end{itemize}

We also added the gradient penalty introduced by~\cite{gulrajani2017improved} to improve the convergence of the GAN loss functions. In summary, the objective for the full end-to-end model is  
\begin{eqnarray}
\mathcal{L}({G}) &=& \lambda_1 [\mathcal{L}_{L_1} (G_\text{c}) + \mathcal{L}_{L_1} (G_\text{dec}) + \mathcal{L}_{L_1} (\text{STN})]\nonumber \\
&+& \lambda_2 \mathcal{L}_\text{CE}(G^M_\text{dec}) \nonumber\\
&+&\lambda_3 [\mathcal{L}_\text{cGAN} (G_{\text{c}},D_{\text{c}}) + \mathcal{L}_\text{cGAN} (G_{\text{dec}},D_{\text{dec}})]\nonumber
\end{eqnarray}

\vspace{-0.3cm}

\section{Experiments}
\label{sec:experiments}
In this section, we study the performance of our compositional GAN model in both the paired and unpaired training regimes through multiple qualitative and quantitative experiments on synthetic and real data sets.
We will present: (1) images generated directly from the composition network, $\hat{c}$, before and after the ESMR step, (2) images generated directly based on the predicted segmentation masks as $\hat{c}_s=\hat{M}_{x} \odot x^T + \hat{M}_{y} \odot y^T$.
In all of our experiments, the training hyper-parameters are $\lambda_1=100$, $\lambda_2 = 50$, $\lambda_3 = 1$, and the inference $\lambda = 100$. Furthermore, similar to~\cite{isola2017image}, we ignore random noise as the input to the generator, and dropout is the only source of randomness in the network.

\subsection{Synthetic data sets }
In this section, we use the Shapenet dataset~\cite{shapenet2015} and study two composition tasks: (1) a chair next to a table, (2) a bottle in a basket. On the chair-table data set, we deal with all four composition challenges, i.e., spatial layout, relative scaling, occlusion, and viewpoint transformation. In the basket-bottle experiment, the main challenge is to predict the correct occluding pixels as well as the relative scaling of the two objects.

\textbf{Composing a chair with a table:}
We manually made a collection of $1\mathrm{K}$ composite images from Shapenet chairs and tables to use as the real joint set, $C$, in the paired and unpaired training schemes.
Chairs and tables in the input-output sets can pose in random azimuth angles in the range $[-180^{\circ}, 180^{\circ}]$ at steps of $10^{\circ}$. As discussed in section~\ref{sec:AFN}, given the segmentation mask of an arbitrary table in a random viewpoint and an input chair, our relative appearance flow network synthesizes the chair in the viewpoint consistent with the table. The synthesized test chairs as $X^{\text{RAFN}}$ are presented in the third row of Figure~\ref{fig:chair-table-basket-bottle}-a.

\textbf{Composing a bottle with a basket:}
We manually composed Shapenet bottles with baskets to prepare a training set of 100 joint examples and trained the model both with and without the paired data.

\subsubsection{Ablation study and baselines}
To study the role of different components of our network, we visualize the predicted outputs at different steps in Figure~\ref{fig:chair-table-basket-bottle}. Trained with either paired or unpaired data, we illustrate output of the generator before and after the ESMR step discussed in section~\ref{sec:finetune}, as $\hat{c}^{\text{before}}$ and $\hat{c}^{\text{after}}$, respectively. The ESMR step sharpens the synthesized images at test time and removes the artifacts generated by the model. Given generated images after being refined and their segmentation masks predicted by our pre-trained mask decomposition network, we also represent outputs as the direct summation of the segments, $\hat{c}_s^{\text{after}}$. Our results from the model trained with unpaired data are comparable with those from paired data. Moreover, we depict the performance of the model without our inpainting network in the eighth row, where occlusions are not correct in multiple examples. More test examples are presented in the Appendix.

In addition, to make sure that our network does not memorize its training samples and can be generalized for each new input test example, we find the nearest neighbor composite example in the training set based on the features of its constituent objects extracted from a pre-trained VGG19 network~\cite{simonyan2014very}. The nearest neighbor examples for each test case are shown in the fourth row of Figure~\ref{fig:chair-table-basket-bottle}. 

In the Appendix, we repeat the experiments with each component of the model removed at a time to study their effect on the final composite image. Moreover, we show the poor performance of two baseline models (CycleGAN~\cite{zhu2017unpaired} and Pix2Pix~\cite{isola2017image}) in the challenging composition task of two input domains in addition to a few failure cases of our model for both paired and unpaired scenarios.

\subsubsection{User evaluations}
We have conducted an Amazon Mechanical Turk (AMT) evaluation~\cite{zhang2016colorful} to compare the performance of our algorithm in different scenarios including training with and without paired data and before and after ESMR. Results from $60$ evaluators are summarized in Table~\ref{AMT-synth}, revealing that even without paired examples during training, our proposed model performs comparably well. In addition, the benefit of the ESMR module in generating higher-quality images is clear from the table.

\begin{table}[t!]
  \caption{AMT user evaluation comparing components of our model on the synthetic datasets. 2nd column: number of test images, 3rd column: $\%$ preferences to after vs. before refinement, 4th column: $\%$ preferences to paired training vs. unpaired.}
  \label{AMT-synth}
  \centering
  \begin{tabular}{lccccc}
    \toprule
  Inputs     & $\#$ test & after-vs-before    & paired-vs-  \\
    &  images &  refinement    & unpaired  \\
    \midrule
    Chair-Table & 90 & $71.3\%$ &  $57\%$ \\
    Basket-Bottle & 45 & $64.2\%$ & $57\%$  \\
    \bottomrule
  \end{tabular}
  \vspace{-4mm}
\end{table}

\begin{figure*}
\centering
\includegraphics[width=\textwidth]{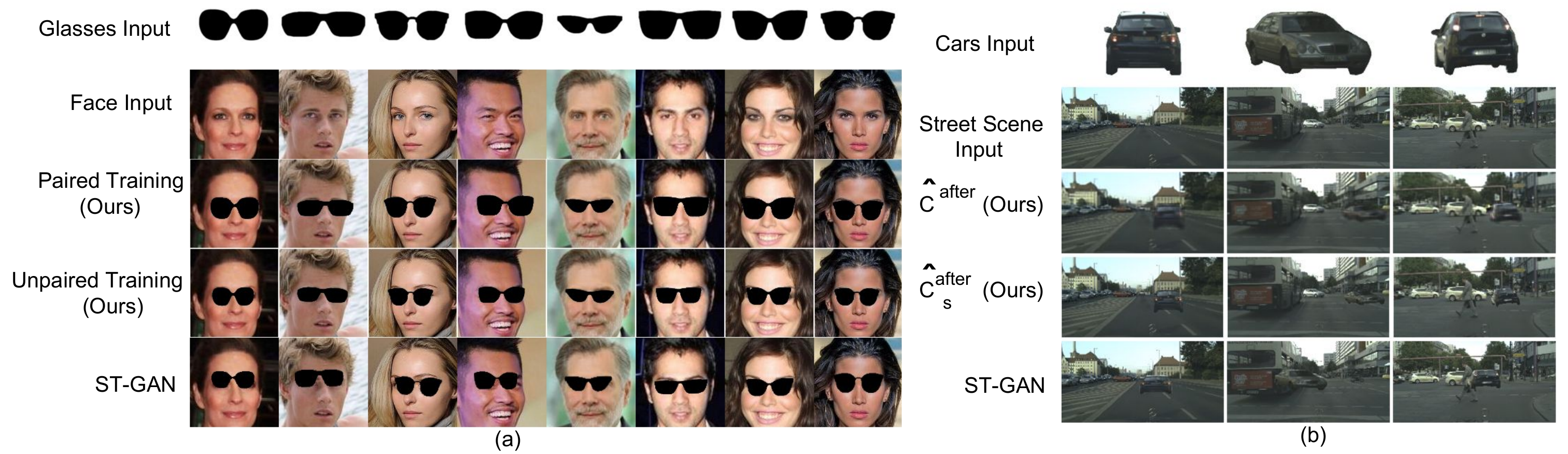}
\caption{(a) Test examples for the face-sunglasses composition task. \textit{Top two rows}: input sunglasses and face images, \textit{3rd and 4th rows}: the output of our compositional GAN for the paired and unpaired models, respectively, \textit{Last row}: images generated by the ST-GAN~\cite{lin2018st} model, (b) Test examples for the street scene-car composition task. \textit{Top two rows}: input cars and street scenes, \textit{3rd and 4th rows}: the output of our compositional GAN after the meta-refinement approach.  Here, $\hat{c}^{\text{after}}$ shows the output of the composition generator and $\hat{c}^{\text{after}}_{\text{s}}$ represents the summation of the masked transposed inputs, \textit{Last row}: images generated by ST-GAN.} 
\vspace{-1mm}
\label{fig:face-sunglasses}
\end{figure*}
\subsection{Real data sets}
 In this section, we use two real datasets to show our model performs equally well when one object is fixed and the other one is relatively scaled and linearly transformed to generate a composed image: (1) a pair of sunglasses to be aligned with a face, (2) a car to be added to a street scene. The problem here is thus similar to the case studies of ST-GAN~\cite{lin2018st} with a background and foreground object.
 
\textbf{Composing a face with sunglasses:}
Here, we used the CelebA dataset~\cite{liu2015faceattributes} and cropped the images to $128\times 128$ pixels. We hand-crafted 180 composite images of celebrity faces from the training split aligned with sunglasses downloaded from the web. In the unpaired scenario, the training set of individual faces in $X$ contains $6\mathrm{K}$ images from the CelebA training split, distinct from the faces in our composite set. Our ST-GAN baseline~\cite{lin2018st} is trained with $\mathrm{10K}$ celebrity faces of the celebA dataset with eyeglasses. 

\textbf{Composing a street scene with a car:}
We used the Cityscapes dataset~\cite{cordts2016cityscapes} and extracted two sets of non-overlapping street scenes from the training set to be used as domains $X$ and $C$ in our compositional setup. In addition, we extracted a set of car images for domain $Y$ using the instance segmentation masks available in the Cityscapes dataset and scaled them to be in a fixed size range. For the scenes collected as our real composite images in set $C$, we use the available segmentation mask of one of the cars in the scene as the mask of the foreground object. We manually filter the real samples that do not include any cars larger than a specific dimension. We also flip images in all three sets for data augmentation. Overall, we have $500$ composite images in set $C$, $1100$ cars images in set $Y$, and $1500$ street scenes in set $X$, all down-scaled to $128\times 256$ pixels. To train the ST-GAN model as a baseline, we used $1600$ Cityscapes street scenes as the real composite images.

\subsubsection{Qualitative analysis and baselines}
In Figure~\ref{fig:face-sunglasses}, we compare the performance of our model with ST-GAN~\cite{lin2018st}, which assumes images of faces (or street scenes) as a fixed background and warps the glasses (or the cars) in a geometric warp parameter space. Our results in the paired and unpaired cases, shown in Figure~\ref{fig:face-sunglasses}, look more realistic in terms of the scale and location of the foreground object. More examples are presented in the Appendix. In the street scene-car composition, we do not have any paired data and can only evaluate the unpaired model. 

\subsubsection{User evaluations}
To confirm our qualitative observations, we asked $60$ evaluators to score our model predictions versus ST-GAN, with the results summarized in Table~\ref{AMT-real}. This experiment confirms the superiority of our network to the state-of-the-art model in composing a background image with a foreground.

\begin{table}[t!]
  \caption{AMT user evaluation comparing our model with ST-GAN on the real datasets. 2nd column: number of test images, 3rd and 4th columns: $\%$ preferences respectively to paired and upaired training vs. ST-GAN. }
  \label{AMT-real}
  \centering
  \begin{tabular}{lccc}
    \toprule
  Inputs     & $\#$ test & paired-vs-  & unpaired-vs-  \\
  &images & ST-GAN & ST-GAN  \\
    \midrule
    Face-Sunglasses & 75 & $84\%$ & $73\%$\\
    Street Scene-Car & 80 & - & $61\%$\\
    \bottomrule
  \end{tabular}
  \vspace{-4mm}
\end{table}

\section{Conclusion and Future Work}
In this paper, we proposed a novel Compositional GAN model addressing the problem of object composition in conditional image generation. Our model captures the relative affine and viewpoint transformations needed to be applied to each input object (in addition to the pixels occlusion ordering) to generate a realistic joint image. We use a decomposition network as a supervisory signal to improve the task of composition both at training and test times. We evaluated our compositional GAN through multiple qualitative experiments and user evaluations for two cases of paired versus unpaired training data on synthetic and real data sets. In the future, we plan to extend this work toward modeling photometric effects (e.g., lighting) in addition to generating images composed of multiple (more than two) and/or non-rigid objects. 


\bibliographystyle{ieee}
\bibliography{egbib}

\newpage
\appendix
\section*{Appendix}

\section{Model Architecture}
In this section, we explain the architecture of our relative appearance flow network and the relative spatial transformer networks in more details.  
\subsection{Relative Appearance Flow Network (RAFN)}
\begin{figure*}[h!]
\centering
\includegraphics[width=0.7\textwidth]{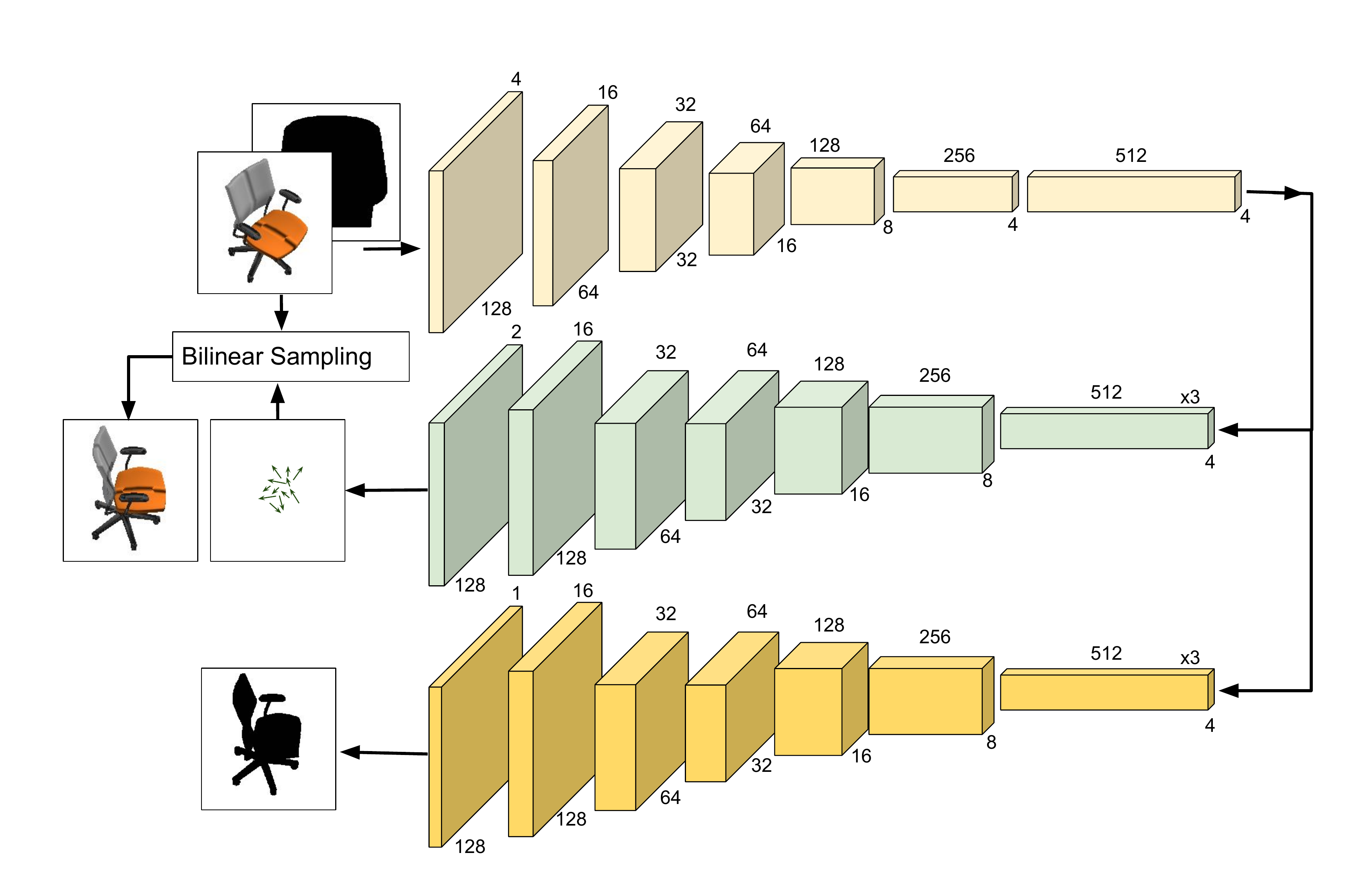}
\caption{Relative Appearance Flow Network: Input is an image of a chair with three RGB channels concatenated channel-wise with the table foreground mask. Output is the appearance flow for synthesizing a new viewpoint of the chair. All layers are convolutional.} 
\label{fig:AFN}
\end{figure*}

Our relative appearance flow network (RAFN) transforms the viewpoint of one object given that of the other object encoded in its binary mask, orthogonal to our main CoDe pipeline. Independent of having access to the paired or unpaired compositional data sets, we need a data set containing different viewpoints of the two objects to supervise this network, as discussed in the paper. The architecture of this network is illustrated in Figure~\ref{fig:AFN}. RAFN is composed of an encoder-decoder set of convolutional layers to predict the appearance flow vectors, which after a bilinear sampling generate the synthesized view of one of the objects consistent with the viewpoint of the other masked input. Another decoder following the same encoder network predicts the foreground mask of the synthesized image, the last row of layers in Figure~\ref{fig:AFN}~\cite{zhou2016view}. All convolutional layers are followed by batch normalization~\cite{ioffe2015batch} and a \texttt{ReLU} activation layer except for the last convolutional layer in each decoder. In the flow decoder, the output is fed into a \texttt{Tanh} layer while in the mask prediction decoder, the last convolutional layer is followed by a \texttt{Sigmoid} to be in the range $[0,1]$. 

\subsection{Relative spatial transformer network}
\label{app:stn}
\begin{figure*}[h!]
\centering
\includegraphics[width=0.7\textwidth]{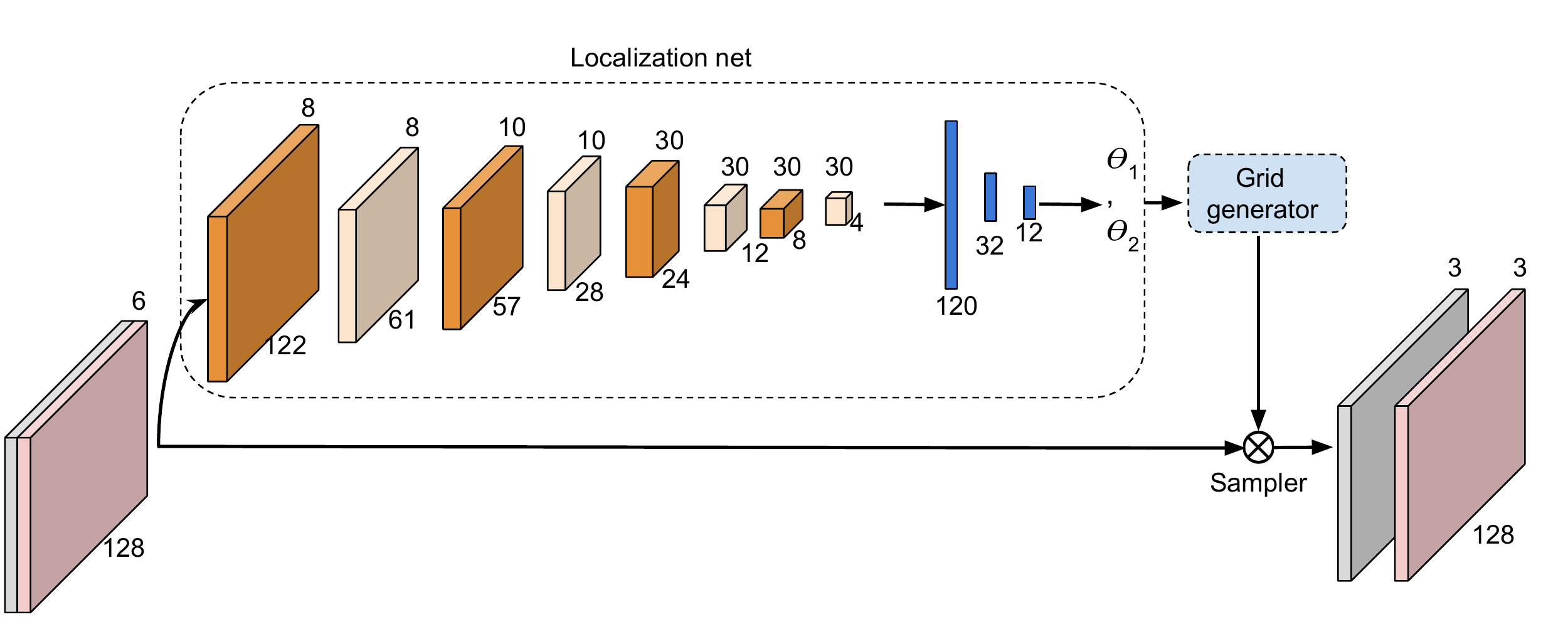}
\caption{Relative Spatial Transformer Network: First input with three RGB channels (e.g., image of a chair) concatenated channel-wise with the second RGB image (e.g., image of a table). The network generates two transformed images each with three RGB channels.} 
\label{fig:STN}
\end{figure*}

The diagram of our relative spatial transformer network is represented in Figure~\ref{fig:STN}. The two input images (e.g., chair and table) are concatenated channel-wise and fed into a localization network~\cite{spatial2015} to generate two sets of parameters, $\theta_1, \theta_2$, for the affine transformations to be applied on each object, respectively. This single network is simultaneously trained on the two images to transform each object to its corresponding transposed target. In this figure, the orange feature maps are the outputs of the \texttt{conv2d} layer (represented along with their corresponding number of channels and dimensions), and the yellow maps are the outputs of the \texttt{max-pool2d} followed by a \texttt{ReLU}. The blue layers also represent fully connected layers.

\section{Additional Results}
Here, we present additional test examples in different compositional domains as well as an ablation study to clarify the role of different components in our proposed model.
\subsection{Composing a chair with a table}
\label{app:fail}
\begin{figure*}[h!]
\centering
\includegraphics[width=0.7\textwidth]{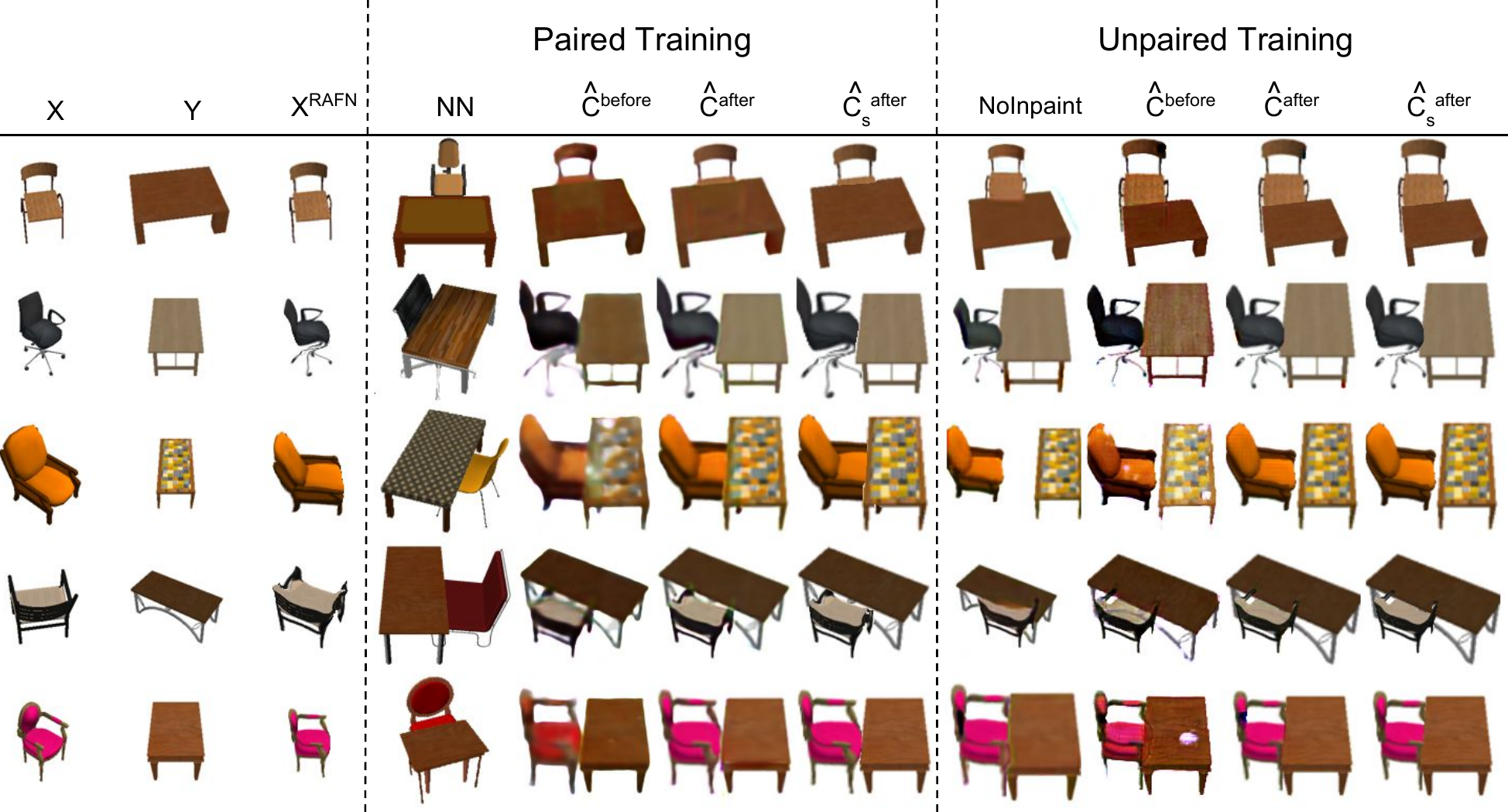}
\caption{Test results on the chair-table composition task trained with either paired  or unpaired data. ``NN'' stands for the nearest neighbor image in the paired training set, and ``NoInpaint'' shows the results of the unpaired model without the inpainting network. In both paired and unpaired cases, $\hat{c}^{\text{before}}$ and $\hat{c}^{\text{after}}$ show outputs of the generator before and after the ESMR approach, respectively. Also, $\hat{c}^{\text{after}}_s$ represents summation of masked transposed inputs after ESMR.} 
\label{fig:chair-table-cont}
\end{figure*}

\begin{figure}
\centering
\includegraphics[width=0.45\textwidth]{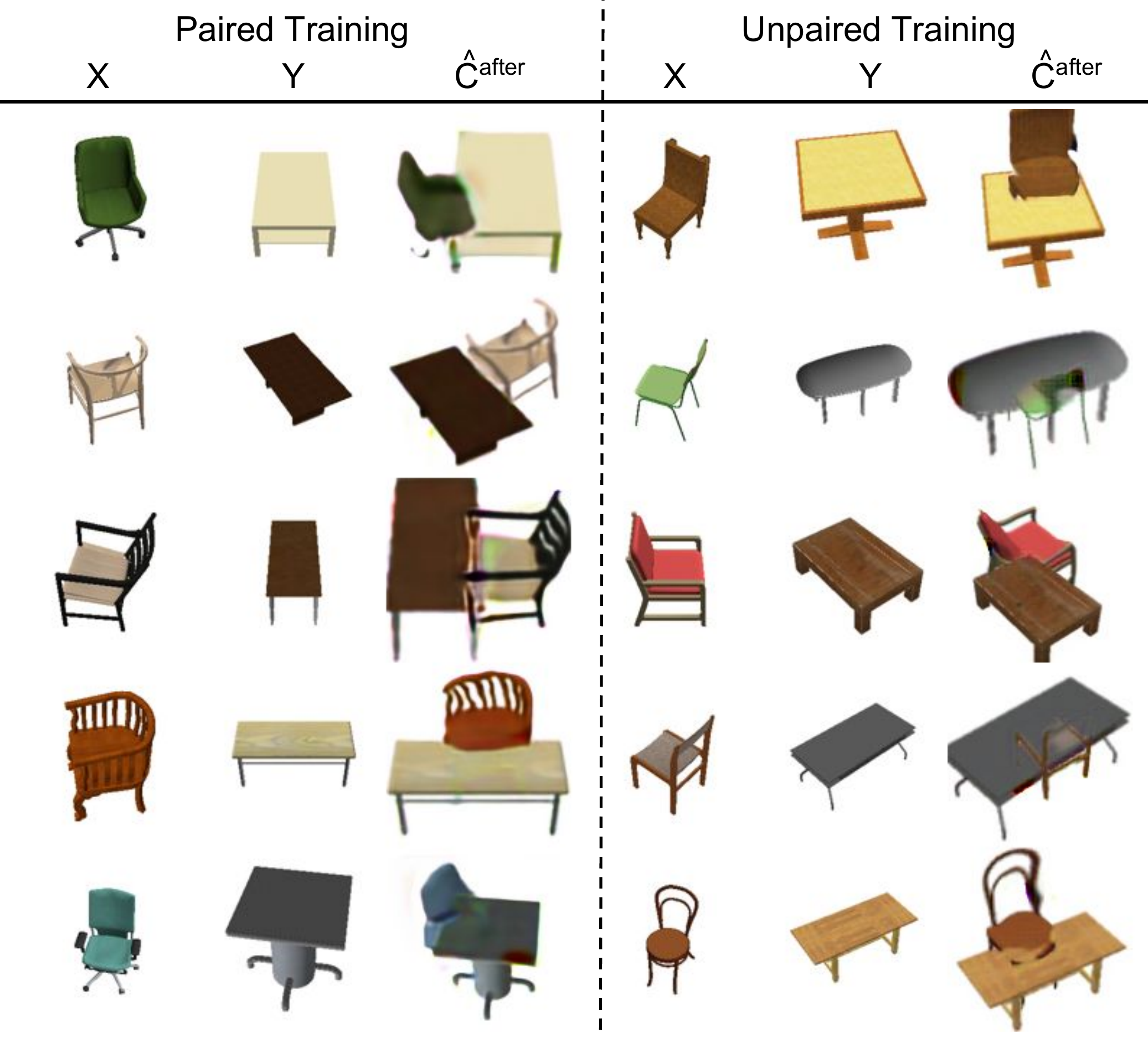}
\caption{Failure test cases for both the paired and unpaired models on the chair-table composition task.} 
\label{fig:chair-table-fail}
\end{figure}

On the challenging problem of composing a chair with a table, we illustrate more test examples in Figure~\ref{fig:chair-table-cont} and a few failure test examples in Figure~\ref{fig:chair-table-fail} for both paired and unpaired training models. Here, the viewpoint and linear transformations in addition to the pixel occlusion ordering should be performed properly to generate a realistic image.

\subsection{Composing a bottle with a basket}
\label{app:basket}
\begin{figure*}[h!]
\centering
\includegraphics[width=0.7\textwidth]{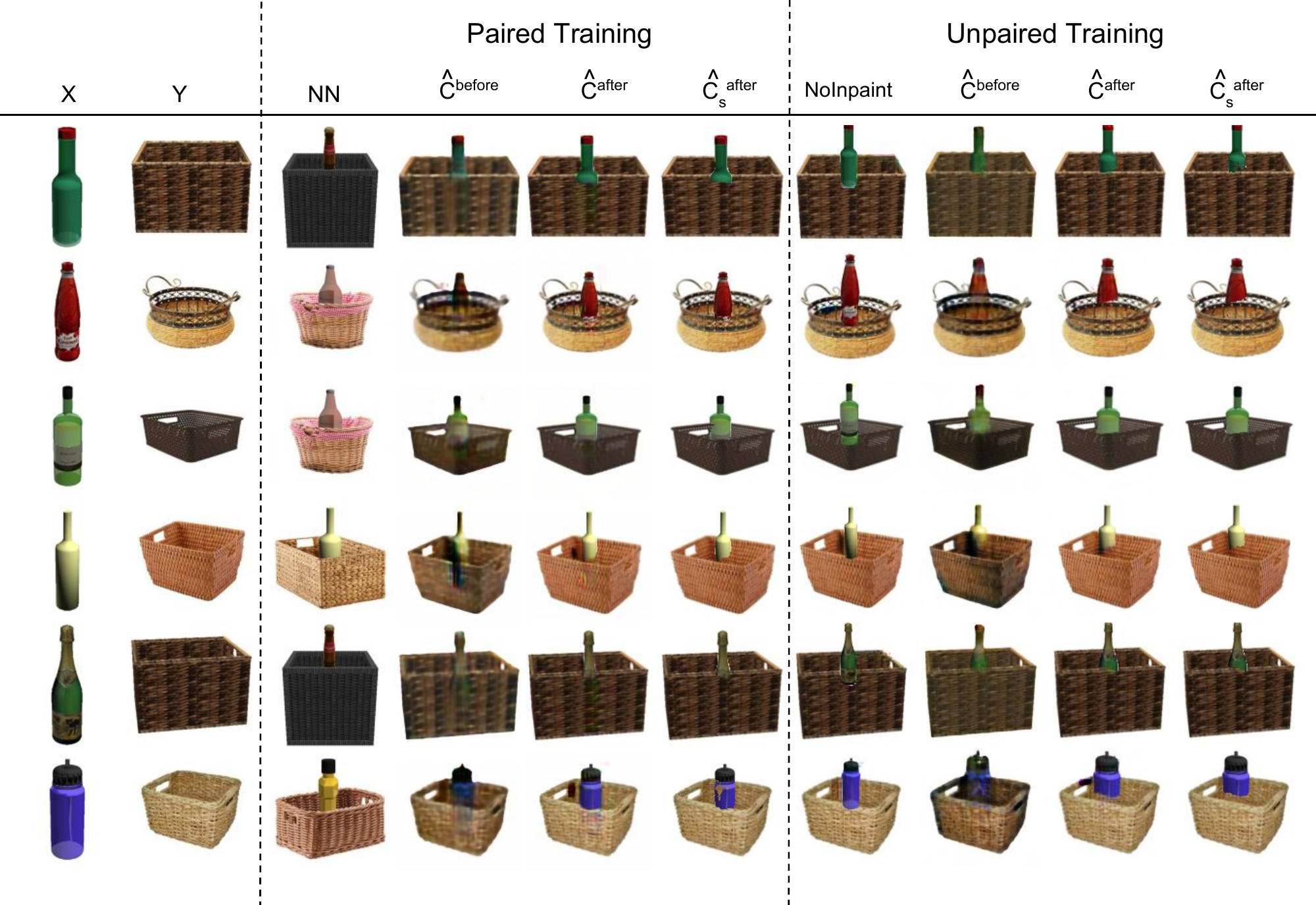}
\caption{More test results on the basket-bottle composition task trained with either paired  or unpaired data. ``NN'' stands for the nearest neighbor image in the paired training set, and ``NoInpaint'' shows the results of the unpaired model without the inpainting network. In both paired and unpaired cases, $\hat{c}^{\text{before}}$ and $\hat{c}^{\text{after}}$ show outputs of the generator before and after the ESMR approach, respectively. Also, $\hat{c}^{\text{after}}_s$ represents summation of masked transposed inputs after ESMR.} 
\label{fig:basket-bottle-cont}
\end{figure*}

In the bottle-basket composition, the main challenging problem is the relative scaling of the objects besides their partial occlusions. In Figure~\ref{fig:basket-bottle-cont}, we visualize more test examples and study the performance of our model before and after the ESMR step for both paired and unpaired scenarios. The third column of this figure represents the nearest neighbor training example found for each new input pair, $(X, Y)$, in terms of their features extracted from the last layer of a pre-trained VGG19 network~\cite{simonyan2014very}. Moreover, the seventh column shows outputs of the network trained with unpaired data when the inpainting component is removed during training. These examples confirm the necessity of the inpainting network while training our compositional GAN model with unpaired data.

\subsection{Ablation study}
\label{app:ablation}
We repeat the experiments on composing a bottle with a basket, with each component of the model removed at a time, to study their effect on the final composite image. Qualitative results are illustrated in Figure~\ref{fig:ablation}. The first and second columns show bottle and basket images concatenated channel-wise as the input of the network. Following columns are:\\
{\textit{- third column}}: no pixel reconstruction loss on the composite image results in the wrong color and faulty occlusion,\\
{\textit{- fourth column}}: no cross-entropy mask loss in training results in faded bottles,\\
{\textit{- fifth column}}: no GAN loss in training and inference generates outputs with a different color and lower quality than the input image, \\
{\textit{- sixth column}}: no decomposition generator ($G_{\text{dec}}$) and self-consistent cycle results in partially missed bottles,\\
{\textit{- seventh, eighth columns}} represent full model in paired and unpaired scenarios, respectively. 

\subsection{Other baselines:}
\label{app:baseline}
\begin{figure*}[t!]
\centering
\includegraphics[width=0.7\textwidth]{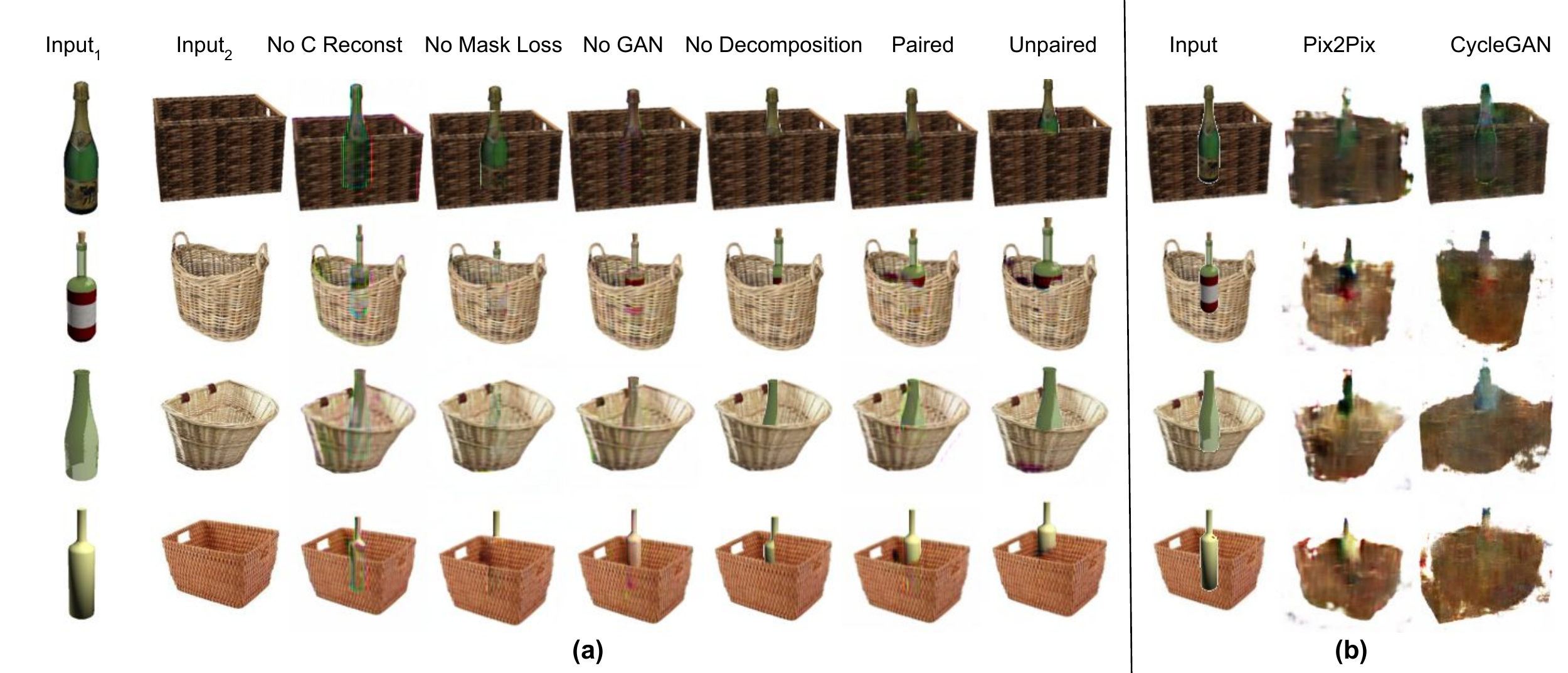}
\caption{(a) Ablation Study: Output of our model without the component specified on top of each column. Input is the channel-wise concatenation of the bottle and basket shown in the first two columns, (b) Baselines: As the input (9th column), each bottle is added to the basket after being scaled and translated with constant parameters. Pix2Pix and CycleGAN outputs are shown on the right.} 
\label{fig:ablation}
\end{figure*}

Our model is designed for the challenging composition problem that requires learning the spatial layout, relative scaling, occlusion, and viewpoint of the two object images to generate a realistic composite image. However, the conditional GAN models such as CycleGAN~\cite{zhu2017unpaired} and Pix2Pix~\cite{isola2017image} address the image translation problem from one domain to another by only changing the appearance of the input image. Here, we compare our model with these two networks in the basket-bottle composition task, where the mean scaling and translating parameters of our training set are used to place each bottle and basket together to have an input with three RGB channels, illustrated in the \textit{ninth column} in Figure~\ref{fig:ablation}. 
We train a ResNet generator on our paired training data with an adversarial loss added with an $L_1$ regularizer. Since the structure of the input image is different from its corresponding ground-truth image, due to different object scalings and layouts, ResNet model works better than a U-Net but still generates unrealistic images, presented in the \textit{tenth column} in Figure~\ref{fig:ablation}. We follow the same approach for the unpaired data and the CycleGAN model with the results in the \textit{eleventh column} in Figure~\ref{fig:ablation}. Our qualitative results confirm the difficulty of learning the transformation between samples from the input distribution and the real composite domain for the Pix2Pix or CycleGAN networks.

\subsection{Composing a face with sunglasses}
\label{app:face}
\begin{figure*}
\centering
\includegraphics[width=0.8\textwidth]{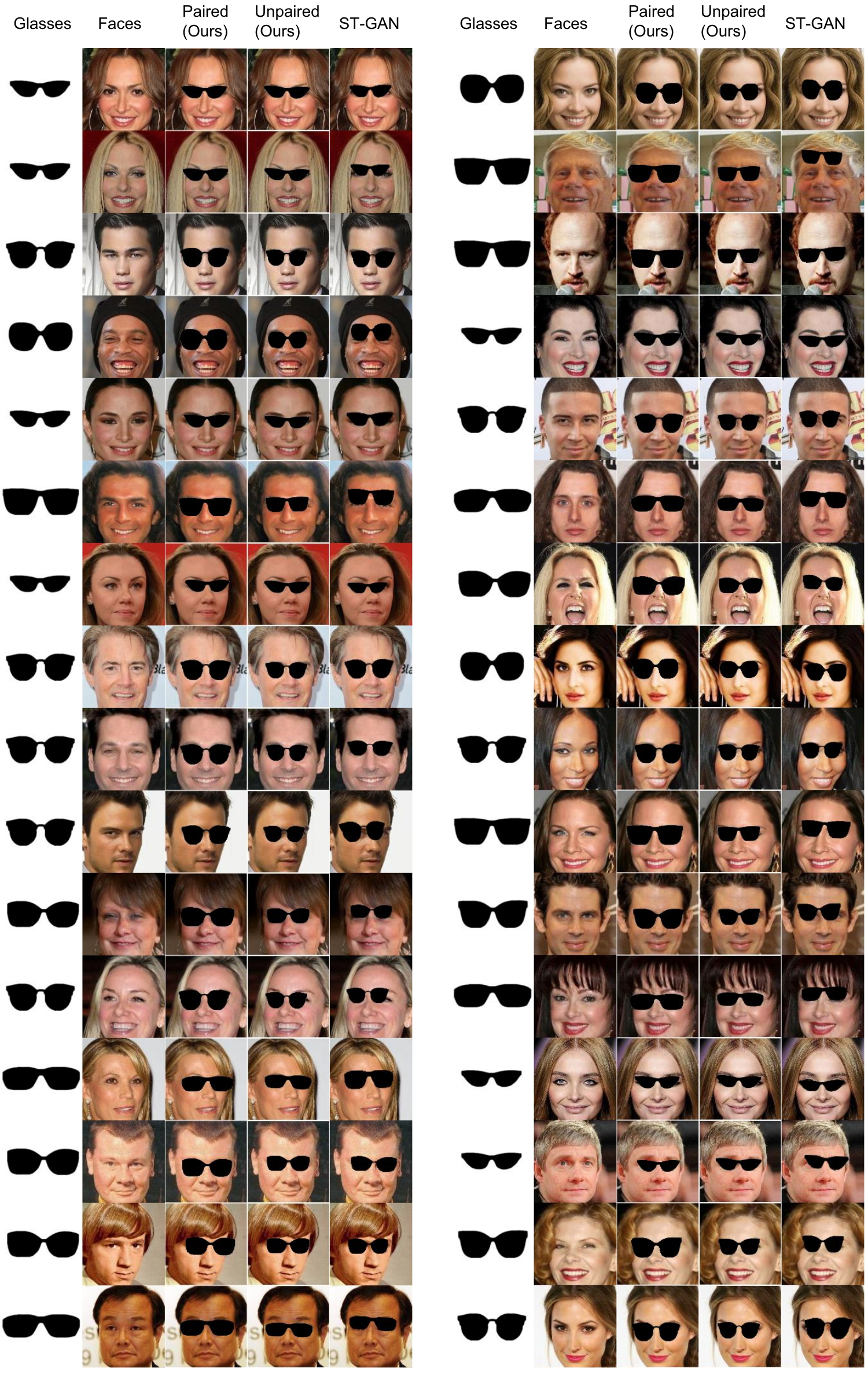}
\caption{Test examples for the face-sunglasses composition task. First two columns show the input sunglasses and face images, 3rd and 4th columns show the output of our compositional GAN for the paired and unpaired models, respectively. Last column shows images generated by the ST-GAN~\cite{lin2018st} model.} 
\label{fig:face-sunglasses-cont}
\end{figure*}

Adding a pair of sunglasses to an arbitrary face image requires a proper linear transformation of the sunglasses to align well with the face. We illustrate several test examples of this composition problem in Figure~\ref{fig:face-sunglasses-cont} including the results of both paired and unpaired training scenarios in the third and fourth columns, respectively. Besides, the last column of each composition example represents the outputs of the ST-GAN model~\cite{lin2018st}.

\subsection{Composing a street scene with a car}
In this section, we illustrate more examples on the composition of the street scenes from the Cityscapes dataset with arbitrary center-aligned cars. This composition requires an appropriate affine transformation to be applied to the cars and make them aligned with the street scene. We compare our results with ST-GAN~\cite{lin2018st} and provide the nearest neighbor training image to each composite test example in terms of their VGG-19 features. 
\begin{figure*}
\centering
\includegraphics[width=\textwidth]{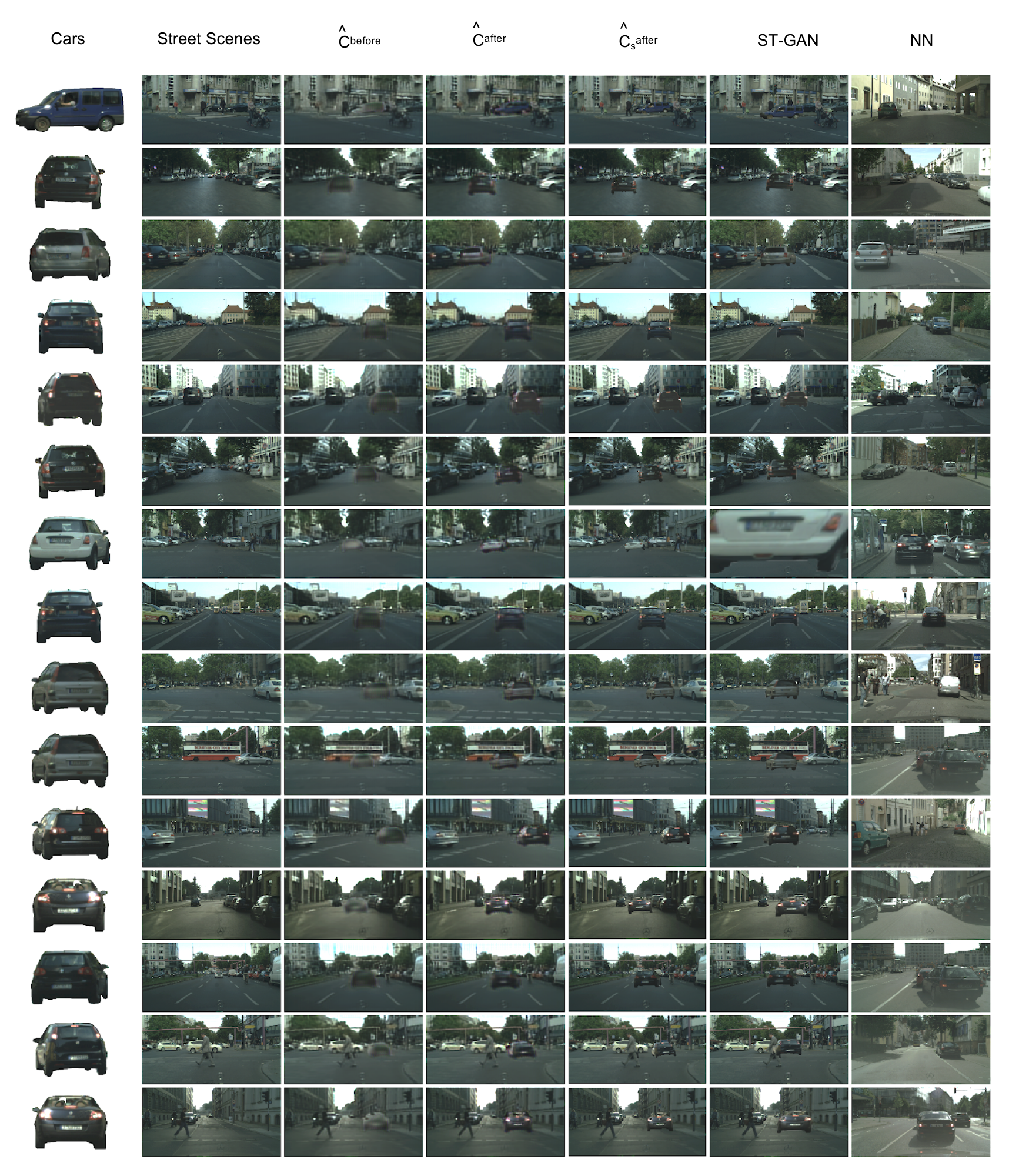}
\caption{Test examples for the street scenes-cars composition task. First two columns show the input car and street images, 3rd and 4th columns show the output of our compositional generator before and after the inference meta-refinement step, respectively. The 5th column shows our model's output by directly adding the masked inputs. The 6th and 7th columns correspond with images generated by the ST-GAN~\cite{lin2018st} model and the nearest neighbor training images.} 
\label{fig:street-car-cont}
\end{figure*}

\end{document}